\pdfoutput=1

\documentclass[11pt]{article}

\usepackage[final]{acl}

\usepackage{times}
\usepackage{latexsym}

\usepackage[T1]{fontenc}

\usepackage[utf8]{inputenc}

\usepackage{microtype}

\usepackage{inconsolata}

\usepackage{graphicx}

\usepackage{natbib}
\usepackage{amsmath}
\usepackage{multirow}
\usepackage{array}
\usepackage{booktabs}
\usepackage{lipsum}
\usepackage[hypcap=true]{caption}
\usepackage{subcaption}
\usepackage{amssymb}
\usepackage{algorithm}
\usepackage{algpseudocode}

\title{QPruner: Probabilistic Decision Quantization for Structured Pruning in Large Language Models}

\author{
 \textbf{Changhai Zhou\textsuperscript{1,3}},
 \textbf{Yuhua Zhou\textsuperscript{2}},
 \textbf{Yibin Wang\textsuperscript{1}},
\\
 \textbf{Shijie Han\textsuperscript{4}},
 \textbf{Qian Qiao\textsuperscript{5}},
 \textbf{Hongguang Li\textsuperscript{3}},
\\
 \textsuperscript{1}Fudan University,
 \textsuperscript{2}Zhejiang University,
 \textsuperscript{3}JF SmartInvest Holdings,
 \textsuperscript{4}Columbia University,
 \textsuperscript{5}Soochow University,
\\
 \small{
   \href{zhouch23@m.fudan.edu.cn}{zhouch23@m.fudan.edu.cn}
 }
 \small{
   \href{zhouyuhua@zju.edu.cn}{zhouyuhua@zju.edu.cn}
 }
 \small{
   \href{yibinwang1121@163.com}{yibinwang1121@163.com}
 }\\
 \small{
   \href{sh4460@columbia.edu}{sh4460@columbia.edu}
 }
 \small{
   \href{qqiao@stu.suda.edu.cn}{qqiao@stu.suda.edu.cn}
 }
 \small{
   \href{harvey2@mail.ustc.edu.cn}{harvey2@mail.ustc.edu.cn}
 }
}

\begin{document}
\maketitle

\begin{abstract}
The rise of large language models (LLMs) has significantly advanced various natural language processing (NLP) tasks. However, the resource demands of these models pose substantial challenges. Structured pruning is an effective approach to reducing model size, but it often results in significant accuracy degradation, necessitating parameter updates to adapt. Unfortunately, such fine-tuning requires substantial memory, which limits its applicability. To address these challenges, we introduce quantization into the structured pruning framework to reduce memory consumption during both fine-tuning and inference. However, the combined errors from pruning and quantization increase the difficulty of fine-tuning, requiring a more refined quantization scheme. To this end, we propose QPruner, a novel framework that employs structured pruning to reduce model size, followed by a layer-wise mixed-precision quantization scheme. Quantization precisions are assigned to each layer based on their importance to the target task, and Bayesian optimization is employed to refine precision allocation strategies, ensuring a balance between model accuracy and memory efficiency. Extensive experiments on benchmark datasets demonstrate that QPruner significantly outperforms existing methods in memory savings while maintaining or improving model performance. 
\end{abstract}


\section{Introduction}

The advent of large language models (LLMs) has revolutionized various natural language processing (NLP) tasks, such as machine translation \cite{zhang2023prompting, sato2020vocabulary}, sentiment analysis \cite{zhang2023enhancing, deng2023llms}, and speech recognition \cite{min2023exploring}. Despite their impressive capabilities, the resource consumption required to obtain a fine-tuned model suitable for specific tasks remains substantial due to the large number of parameters and high computational demands of LLMs \cite{frantar2023sparsegpt}. To address these issues, various compression techniques, including pruning \cite{molchanov2019importance, liu2018rethinking}, quantization \cite{shao2023omniquant, lee2023enhancing}, and distillation \cite{gu2023minillm, tan2023gkd}, have been proposed. 

Structured pruning \cite{ma2023llm, xia2023sheared} is a widely used approach that reduces model size by removing less important parameters in a structured manner, preserving the overall architecture compatibility with hardware requirements. However, the disruption of computational graph uniformity and the removal of parameters can significantly reduce the accuracy of LLMs, which are inherently information-dense networks. To mitigate this degradation, fine-tuning is often used to recover the accuracy of pruned models. This fine-tuning step, while effective, is memory-intensive and presents substantial challenges in terms of resource consumption.

To further reduce memory usage during the fine-tuning and inference phases, we introduce quantization into the structured pruning framework. Specifically, after performing structured pruning, we quantize the pruned model and then apply different fine-tuning strategies. Quantization effectively reduces the bit-width of model parameters, thereby lowering the resource consumption during both fine-tuning and inference. However, integrating quantization with structured pruning introduces additional complexities. Structured pruning applies different pruning intensities across model layers, which exacerbates the uneven distribution of layer importance, making some layers more critical for maintaining model performance. Moreover, the cumulative quantization error varies across different layers, potentially amplifying the performance degradation caused by pruning. Therefore, a simple, uniform quantization scheme is suboptimal. Instead, a more nuanced, layer-wise mixed-precision quantization approach is needed. By allowing more critical layers to maintain higher precision, we can better control the overall performance of the model.

Building upon these observations, we propose a new framework called QPruner. In QPruner, we first apply structured pruning to reduce the model size, followed by a quantization phase where different quantization precisions are assigned to each layer based on their contribution to the target task. To further improve the allocation strategy, Bayesian optimization \cite{frazier2018bayesian} is employed to explore better precision configurations. Finally, we apply parameter-efficient fine-tuning (PEFT) fine-tuning strategy, to recover model performance. This integrated approach aims to strike an optimal balance between model accuracy and memory efficiency, making it well-suited for resource-constrained scenarios. The main contributions of this work are summarized as follows:
\begin{itemize}
    \item We propose QPruner, a novel framework that integrates structured pruning and quantization, aiming to significantly reduce the memory consumption of LLMs during both fine-tuning and inference.
    \item We introduce a mixed-precision quantization scheme where quantization precisions are assigned to each layer based on their importance to the target task, with Bayesian optimization used to further refine precision allocation strategies.
    \item We demonstrate QPruner's powerful ability to save memory and maintain performance. It can surpass baseline methods in terms of accuracy by up to 6\% while saving at least 30\% of memory.
\end{itemize}


\section{Background and Motivation}

\subsection{Quantization}
\textbf{Quantization.} Quantization is an essential technique used to reduce the computational and memory overhead of large-scale models by converting high-precision numerical values, such as a 32-bit floating-point number $X^{\text{HP}} \in \mathbb{R}$, into a lower-bit integer representation $X^{\text{INT}} \in \{0, 1, \dots, 2^{N}-1\}$. This process is mathematically expressed as:
\begin{align}
\label{eq:quant}
    X^{\text{INT}} = \text{round}\left((2^{N} - 1) F\left( X^{\text{HP}} \right) \right),
\end{align}
where $F(\cdot) \colon \mathbb{R} \rightarrow [0, 1]$ is a normalization function. A typical method is uniform quantization, where $F(X)$ is defined as $F(X) = \frac{X - X_{\min}}{X_{\max} - X_{\min}}$. An alternative approach introduced by QLoRA \citet{dettmers2024qlora} is 4-bit NormalFloat Quantization (NF4), which assumes that the data follows a normal distribution $X \sim \mathcal{N}(0, \sigma^2)$ and applies $F(X) = \Phi(X/\sigma)$, with $\Phi(\cdot)$ representing the cumulative distribution function of a standard normal distribution.

\textbf{Dequantization.} To recover the high-precision values from their quantized forms, a lookup table $\mathcal{T}$ is used, which is defined as:
\begin{align}
\label{eq:lookup_table}
    \mathcal{T}[i] = F^{-1}\left( \frac{i}{2^N - 1} \right), \quad i = 0, 1, \dots, 2^{N}-1,
\end{align}
allowing the integer $X^{\text{INT}}$ to be mapped back to its simulated high-precision counterpart $X^{\text{D}} \in \mathbb{R}$. The dequantization process can be represented as:
\begin{align}
\label{eq:dequant}
    X^{\text{D}} = \mathcal{T}[X^{\text{INT}}].
\end{align}

\textbf{Simulated Quantization for Matrices.} In practice, it is often more efficient to use simulated quantization for matrices rather than directly operating on quantized values \citep{bai2020binarybert, shen2020q}. In this method, quantized weight matrices are stored as encoded integers and are temporarily dequantized into simulated high-precision matrices during multiplication operations. This process is denoted by $q_N(\cdot) \colon \mathbb{R}^{m \times n} \rightarrow \mathbb{R}_{N}^{m \times n}$, where $\mathbb{R}_{N}: \{\mathcal{T}[i] \in \mathbb{R} | 0 \leq i < 2^N \}$.

\begin{figure*}[t]
    \centering
    \includegraphics[width=0.85\textwidth]{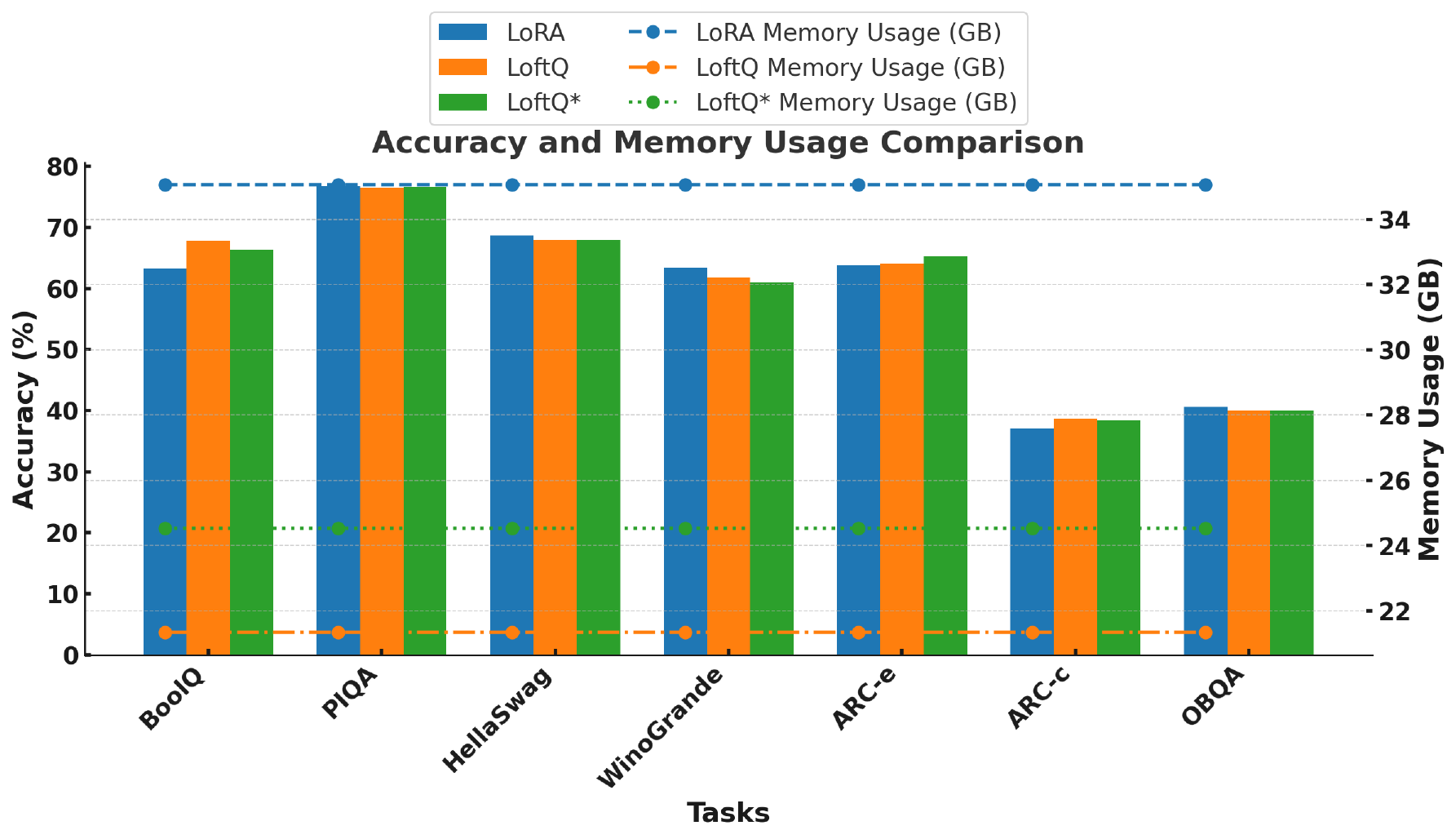}
    \caption{Comparison of accuracy and memory usage across different fine-tuning configurations for multiple tasks. The bars represent the accuracy of three different methods (LoRA, LoftQ, LoftQ*) on each task, while the markers indicate the memory usage for each corresponding method.}
    \label{fig:moti}
\end{figure*}

\subsection{The Motivating Example}\label{motivation}

Efficient fine-tuning of LLMs on resource-constrained devices requires effective model compression and fine-tuning techniques. After applying structured pruning and quantization, more efficient fine-tuning methods are needed to recover accuracy. One approach is to use LoRA-based methods, as done in LLM-Pruner \cite{ma2023llm}, which employs LoRA for quick recovery after structured pruning. Among the LoRA series methods, LoftQ \citet{li2023loftq} is a method for fine-tuning quantized models. Before fine-tuning, LoftQ iteratively updates the low-rank matrices such that the quantized matrix $\mathbf{Q} + \mathbf{AB}$ approximates the full-precision matrix $\mathbf{W}$, thereby improving the fine-tuning performance, particularly in low-bit settings.

Simply combining pruning, quantization, and LoRA can lead to suboptimal results. Structural pruning reduces model size by removing less important parameters, but due to the varying importance of different layers, it often results in uneven pruning across layers. This uneven pruning leads to a complex and unbalanced network structure, and standard quantization typically applies a uniform configuration across all layers. To explore a better trade-off between performance and memory, we adopted mixed-precision quantization, assigning different computational resources and complexities to different layers, with the goal of allowing more important layers to learn with finer granularity.

We conducted experiments using the LLaMA-7b model with a pruning rate of 20\%. The pruning was performed using the optimal strategy determined by LLM-Pruner. The methods compared were as follows: LoRA with a uniform 16-bit configuration, LoftQ with a uniform 4-bit quantization, and LoftQ* with a mixed-precision setting of 4 or 8 bits per layer. As shown in Figure \ref{fig:moti}, the quantized models (LoftQ) achieved performance comparable to the original precision models (LoRA), with significantly lower memory usage (21.33 GB versus 35.06 GB). On some tasks, there was a slight drop in performance, but the mixed-precision model (LoftQ*) demonstrated the potential to further enhance performance while maintaining efficient memory usage.
\section{QPruner}

\begin{figure*}[t]
    \centering
    \includegraphics[width=\textwidth]{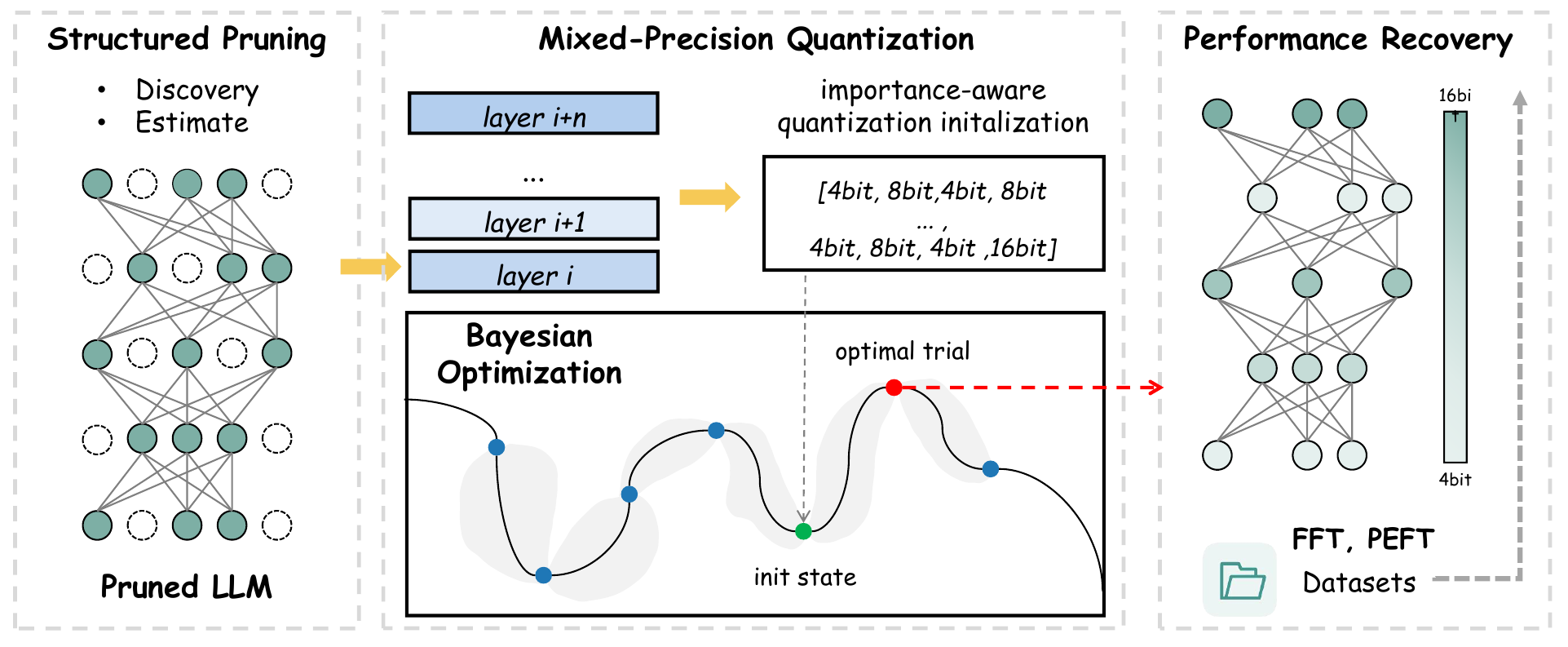}
    \caption{Overview of the QPruner framework.}
    \label{fig:over}
\end{figure*}

Structured pruning, while effective in reducing model size, can disrupt the balance of layer importance, leading to performance degradation. Therefore, parameter adjustments are often necessary to mitigate this imbalance and restore model performance. However, parameter updates require significant memory, which is why we employ quantization techniques to reduce memory consumption. As demonstrated in the motivating example, simply combining pruning and quantization is not always the best choice, as the importance of different layers in a pruned model can vary greatly. We need finer-grained layer-wise quantization bit-width control, which introduces a challenging bit-width allocation problem. To address this, we designed a two-stage allocation strategy to effectively balance these trade-offs.

Based on these insights, we propose QPruner, an integrated framework tailored for efficient or low-resource NLP tasks. It employs structured pruning, mixed-precision quantization, and efficient fine-tuning to solve the challenges of balancing memory efficiency and model performance.

\subsection{Structured Pruning}\label{3.1}
Our framework does not impose specific requirements on the pruning method; as new technologies evolve, the pruning method can be replaced. The only requirement for this step is to produce a smaller model. Although some methods can achieve good performance without fine-tuning \cite{an2024fluctuation}, most real-time systems require dynamic adaptation, which means that the pruned model must be fine-tuned to improve performance.

A popular structured pruning method is LLM-Pruner \cite{ma2023llm}, which first identifies dependencies between neurons and groups them, then removes weights based on their importance. Let $N_i$ and $N_j$ be two neurons in the model. If $N_j \in \text{Out}(N_i)$ and $\text{Deg}^-(N_j) = 1$, then $N_j$ is dependent on $N_i$. Similarly, if $N_i \in \text{In}(N_j)$ and $\text{Deg}^+(N_i) = 1$, then $N_i$ is dependent on $N_j$. Based on this principle, a dependency graph can be constructed to iteratively identify all coupled structures.

Next, these coupled structures are grouped, and their importance is estimated to effectively perform pruning. For a group of coupled structures $\mathbf{G} = \{\mathbf{W}_i\}_{i=1}^M$, its importance can be expressed as:

\begin{equation}
I_{\mathbf{W}_i} = |\mathcal{L}_{\mathbf{W}_i}(\mathcal{D}) - \mathcal{L}_{\mathbf{W}_i=0}(\mathcal{D})|,
\end{equation}

where $\mathcal{L}$ represents the prediction loss.

Using a second-order Taylor expansion, the importance can be approximated as:

\begin{equation}
\left| \frac{\partial \mathcal{L}(\mathcal{D})}{\partial \mathbf{W}_i} \mathbf{W}_i - \frac{1}{2} \mathbf{W}_i^\top \mathbf{H} \mathbf{W}_i \right|,
\end{equation}

where $\mathbf{H}$ is the Hessian matrix of the loss function.

For each parameter $W_k^i$, its importance is defined as:

\begin{equation}
\left| \frac{\partial \mathcal{L}(\mathcal{D})}{\partial W_k^i} W_k^i - \frac{1}{2} (W_k^i)^2 H_{kk} \right|,
\end{equation}

where $H_{kk}$ is the $k$-th diagonal element of the Hessian matrix.

Finally, we aggregate the importance of each structure into group-level importance using methods such as summation, multiplication, taking the maximum, or using only the last item. Groups with the lowest importance are selected for pruning, thereby reducing the model size while maintaining performance as much as possible.

\subsection{Mixed-Precision Quantization}\label{3.2}
After pruning, we apply mixed-precision quantization to further reduce memory usage while maintaining model performance. Instead of assigning a uniform bit-width across all layers, different bit-widths are allocated based on each layer’s contribution to the final model output. The contribution of each layer is quantified using mutual information between the layer’s output and the model’s prediction.

To compute mutual information, we first run representative data samples through the pruned model. For each layer, we record its output $X$ and the final prediction $Y$. The mutual information $I(X; Y)$ between the output of layer $X$ and prediction $Y$ is computed as:

\begin{equation}
I(X; Y) = \sum_{x \in X} \sum_{y \in Y} p(x, y) \log \frac{p(x, y)}{p(x)p(y)},
\end{equation}

where $p(x, y)$ is the joint probability distribution of $X$ and $Y$, while $p(x)$ and $p(y)$ are the marginal distributions. A higher mutual information value indicates that the layer is more important for the final output and should therefore be assigned a higher bit-width.
\begin{algorithm}
\caption{Mixed-Precision Quantization}
\begin{algorithmic}
\State Compute mutual information $I(X_i; Y)$
\State Initialize bit-width configuration $\mathbf{b}_0$ based on $I(X_i; Y)$ and memory constraint
\State $\mathcal{D} \gets \left\{ (\mathbf{b}_0, P(\mathbf{b}_0), M(\mathbf{b}_0)) \right\}$ 
\While{not converged}
    \State Train GP model on $\mathcal{D}$
    \State $\mathbf{b}_{t+1} \gets \arg\max_{\mathbf{b}} \alpha(\mathbf{b})$
    \State Apply $\mathbf{b}_{t+1}$ to pruned model and fine-tune
    \State Measure $P(\mathbf{b}_{t+1})$, $M(\mathbf{b}_{t+1})$
    \State $\mathcal{D} \gets \mathcal{D} \cup \left\{ (\mathbf{b}_{t+1}, P(\mathbf{b}_{t+1}), M(\mathbf{b}_{t+1})) \right\}$
\EndWhile
\end{algorithmic}
\end{algorithm}
Once the mutual information is computed, an average bit-width $B_{\text{avg}}$ is determined based on the available memory budget. Layers with higher importance receive more bits, and the allocation is performed in discrete bit-widths (e.g., 4-bit, 8-bit), constrained by the total memory limit.

Although the initial bit-width configuration derived from mutual information offers a reasonable starting point for fine-tuning, the complex interactions between layers, particularly in LLMs, mean that the importance of individual layers may shift after fine-tuning. As a result, the initial bit-width assignment might not represent the optimal configuration. To further refine the precision configuration, we employ Bayesian optimization.

The objective of Bayesian optimization is to maximize model performance while minimizing memory usage. Let $\mathbf{b} = [B_1, B_2, \dots, B_L]$ represent the bit-width configuration across $L$ layers. The optimization problem is formulated as:

\begin{equation}
\mathbf{b}_{\text{opt}} = \arg\max_{\mathbf{b}} \ \alpha(\mathbf{b}),
\end{equation}

where $\alpha(\mathbf{b})$ is an acquisition function that balances exploration (of less well-understood configurations) and exploitation (of known promising configurations). The memory usage $M(\mathbf{b})$ is constrained by $M_{\text{max}}$, the total available memory.

The process starts by initializing a dataset $\mathcal{D}$ with the initial bit-width configuration $\mathbf{b}_0$, along with its corresponding performance $P(\mathbf{b}_0)$ and memory usage $M(\mathbf{b}_0)$. A Gaussian Process (GP) model is then trained on the data to predict model performance and the uncertainty for new configurations. Based on this model, the acquisition function $\alpha(\mathbf{b})$ is used to select the next bit-width configuration to evaluate. 

Once a new configuration $\mathbf{b}_{t+1}$ is selected, it is applied to the pruned model, fine-tuned, and its performance $P(\mathbf{b}_{t+1})$ and memory usage $M(\mathbf{b}_{t+1})$ are measured. These results are then added to the dataset $\mathcal{D}$, and the GP model is updated with the new data. This iterative process continues until a stopping criterion is met, such as convergence or a maximum number of iterations. Over time, this method refines the bit-width configuration to achieve an optimal balance between model performance and memory efficiency.

\subsection{Performance Recovery}\label{3.3}

After the steps of structured pruning and mixed-precision quantization, significant memory savings are achieved. However, model performance typically needs to be restored through fine-tuning. Full-parameter fine-tuning is often impractical due to the large memory footprint it requires, but our compression technique makes full model fine-tuning feasible by reducing both memory and computational costs.

In addition to traditional full-parameter fine-tuning, efficient fine-tuning techniques such as LoRA (Low-Rank Adaptation) \cite{hu2021lora} have proven especially effective, particularly in scenarios with limited data. LoRA significantly reduces the number of trainable parameters by freezing the original weight matrix $\mathbf{W}_0$ and only updating the low-rank approximation of the weight matrix, represented as $\Delta \mathbf{W} = \mathbf{A} \mathbf{B}$, where $\mathbf{A} \in \mathbb{R}^{d \times r}$ and $\mathbf{B} \in \mathbb{R}^{r \times d}$. Here, $r$ (the rank) is much smaller than the original dimension $d$, leading to a substantial reduction in the number of trainable parameters.

The forward computation in this approach can be written as:

\begin{equation}
\mathbf{Y} = \mathbf{W}_0 \mathbf{X} + \Delta \mathbf{W} \mathbf{X} = \mathbf{W}_0 \mathbf{X} + \mathbf{A} \mathbf{B} \mathbf{X},
\end{equation}

There are also LoRA-like methods specifically designed for quantized models, such as QLoRA \cite{dettmers2023qlora} and LoftQ \cite{li2023loftq}. LoftQ iteratively updates the low-rank matrices $\mathbf{A}$ and $\mathbf{B}$ such that the quantized matrix $\mathbf{Q} + \mathbf{A} \mathbf{B}$ approximates the original full-precision matrix $\mathbf{W}$ during fine-tuning. The objective is defined as:

\begin{equation}
\min_{\mathbf{A}, \mathbf{B}} \| \mathbf{W} - (\mathbf{Q} + \mathbf{A} \mathbf{B}) \|^2.
\end{equation}

where $\mathbf{Q}$ is the quantized matrix.

By combining structured pruning, mixed-precision quantization, and performance recovery techniques, QPruner is able to achieve robust adaptability with minimal computational overhead.

\begin{table*}[t]
    \centering
    \caption{Zero-shot performance and peak memory usage on LLaMA-7B and Vicuna-7B with varying pruning rates. LLM-Pruner represents the currently widely used half-precision model. The performance is reported in percentage (\%), and the memory usage is in gigabytes (GB).}
    \label{table:result-comp}
    \resizebox{\textwidth}{!}{
    \begin{tabular}{@{}llccccccccc@{}}
        \toprule
        & & Method & BoolQ & PIQA & HellS & WinoG & ARC-e & ARC-c & OBQA & Memory (GB) \\
        \midrule
        \multirow{10}{*}{\parbox{2cm}{LLaMA-7B}} & \multirow{1}{*}{Rate = 0\%} 
         & w/o tuning & 73.09 & 78.35 & 72.98 & 67.09 & 67.42 & 41.38 & 42.40 & - \\
        \cmidrule{2-11}
         & \multirow{4}{*}{Rate = 20\%} 
         & LLM-Pruner  & 63.30 & 76.82 & 68.68 & \textbf{63.38} & 63.76 & 37.11 & 40.60 & 35.06 \\
         & & QPruner\textsuperscript{1} & 67.77 & 76.55 & 68.03 & 61.80 & 64.06 & 38.65 & 40.00 & 21.78 \\
         & & QPruner\textsuperscript{2} & 68.60 & 76.79 & 68.43 & 62.78 & 65.50 & 38.74 & 40.40 & 23.05 \\
         & & QPruner\textsuperscript{3} & \textbf{69.11} & \textbf{77.23} & \textbf{68.80} & 63.17 & \textbf{66.16} & \textbf{39.20} & \textbf{41.00} & 23.32 \\
        \cmidrule{2-11}
         & \multirow{4}{*}{Rate = 30\%} 
         & LLM-Pruner  & 62.45 & 74.37 & \textbf{63.14} & \textbf{61.96} & \textbf{59.22} & 33.70 & \textbf{39.60} & 31.38 \\
         & & QPruner\textsuperscript{1} & 58.96 & 71.22 & 58.10 & 58.88 & 52.19 & 32.34 & 38.40 & 20.12 \\
         & & QPruner\textsuperscript{2} & 62.20 & 72.88 & 60.64 & 60.50 & 55.61 & 33.56 & 38.40 & 22.87 \\
         & & QPruner\textsuperscript{3} & \textbf{66.50} & \textbf{74.43} & 61.14 & 61.40 & 58.12 & \textbf{34.47} & 39.20 & 22.15 \\
        \cmidrule{2-11}
         & \multirow{4}{*}{Rate = 50\%} 
         & LLM-Pruner  & 43.76 & 68.88 & 44.85 & 50.99 & 45.20 & 28.75 & 34.60 & 23.89 \\
         & & QPruner\textsuperscript{1} & 45.14 & 68.34 & 44.39 & 52.96 & 43.86 & 29.01 & 35.80 & 15.47 \\
         & & QPruner\textsuperscript{2} & 47.08 & 68.85 & 45.53 & 53.65 & 44.31 & 29.36 & 36.20 & 16.85 \\
         & & QPruner\textsuperscript{3} & \textbf{48.37} & \textbf{69.20} & \textbf{45.19} & \textbf{54.45} & \textbf{45.28} & \textbf{29.70} & \textbf{36.40} & 16.65 \\
        \midrule
        \multirow{10}{*}{\parbox{2cm}{Vicuna-7B}} & \multirow{1}{*}{Rate = 0\%} 
         & w/o tuning & 75.69 & 77.75 & 71.06 & 67.80 & 69.07 & 40.78 & 42.20 & - \\
        \cmidrule{2-11}
         & \multirow{4}{*}{Rate = 20\%} 
         & LLM-Pruner  & 57.77 & 77.56 & 67.16 & 63.14 & 67.30 & 37.71 & 40.40 & 35.25 \\
         & & QPruner\textsuperscript{1} & 57.95 & 76.82 & 66.42 & 62.51 & 66.62 & 37.37 & 40.60 & 21.65 \\
         & & QPruner\textsuperscript{2} & 59.70 & 77.20 & 66.31 & 62.66 & 67.12 & 37.48 & 40.80 & 22.95 \\
         & & QPruner\textsuperscript{3} & \textbf{59.85} & \textbf{77.59} & \textbf{67.31} & \textbf{63.20} & \textbf{67.84} & \textbf{37.85} & \textbf{41.20} & 23.10 \\
        \cmidrule{2-11}
         & \multirow{4}{*}{Rate = 30\%} 
         & LLM-Pruner  & \textbf{58.81} & 74.37 & 60.70 & \textbf{60.62} & 59.01 & 33.79 & 38.80 & 31.83 \\
         & & QPruner\textsuperscript{1} & 53.85 & 74.76 & 60.65 & 60.06 & 59.72 & 34.30 & 38.20 & 19.95 \\
         & & QPruner\textsuperscript{2} & 55.64 & 75.07 & 61.65 & 60.31 & 59.54 & 34.47 & 38.60 & 21.65 \\
         & & QPruner\textsuperscript{3} & 57.23 & \textbf{75.90} & \textbf{62.00} & 60.37 & \textbf{60.81} & \textbf{34.79} & \textbf{39.40} & 21.80 \\
        \cmidrule{2-11}
         & \multirow{4}{*}{Rate = 50\%} 
         & LLM-Pruner  & 59.51 & 66.87 & 43.18 & 52.01 & 48.40 & 26.45 & 34.00 & 24.55 \\
         & & QPruner\textsuperscript{1} & 59.51 & 67.90 & 43.30 & 50.83 & 48.82 & 27.49 & 34.60 & 14.50 \\
         & & QPruner\textsuperscript{2} & 61.31 & 68.56 & 44.54 & 53.02 & 49.50 & 28.13 & 35.40 & 15.90 \\
         & & QPruner\textsuperscript{3} & \textbf{61.56} & \textbf{68.80} & \textbf{43.72} & \textbf{53.39} & \textbf{49.66} & \textbf{27.98} & \textbf{35.80} & 15.35 \\
        \bottomrule
    \end{tabular}}
\end{table*}

\section{Experiments}
\label{4.1}
\begin{table*}[t]
\centering
\caption{Performance comparison (\%) of ablation studies on seven tasks at 20\% pruning rate on LLaMA-7B. It appears that QPruner captures potential resource allocations without relying on other settings.}
\resizebox{\linewidth}{!}{
\begin{tabular}{l|cc|ccc|ccc|cc}
\toprule
\multirow{2}{*}{\textbf{Benchmark}} & \multicolumn{2}{c|}{\textbf{Dtype of 4-bit}} & \multicolumn{3}{c|}{\textbf{Adapter Initialization Method}} & \multicolumn{3}{c|}{\textbf{Adapter Iteration Count}} & \multicolumn{2}{c}{\textbf{Importance Estimation}} \\
\cmidrule(lr){2-3} \cmidrule(lr){4-6} \cmidrule(lr){7-9} \cmidrule(lr){10-11}
 & \textbf{NF4} & \textbf{FP4} & \textbf{LoftQ} & \textbf{Gaussian} & \textbf{PiSSA} & \textbf{iter=1} & \textbf{iter=2} & \textbf{iter=4} & \textbf{Element\textsuperscript{1}} & \textbf{Element\textsuperscript{2}} \\
\midrule
ARC-e       & \textbf{65.49} & 62.84 & \textbf{65.49} & 64.77 & 64.44 & \textbf{65.49} & 64.31 & 64.18 & \textbf{65.49} & 62.50 \\
ARC-c       & \textbf{38.99} & 36.77 & \textbf{38.99} & \textbf{38.99} & 38.40 & \textbf{38.99} & 38.05 & 38.14 & \textbf{38.99} & 37.80 \\
WinoGrande  & 61.40 & \textbf{63.22} & 61.40 & \textbf{61.96} & 61.48 & \textbf{61.40} & 60.46 & 60.69 & \textbf{61.40} & 59.43 \\
OBQA        & \textbf{40.20} & 39.80 & 40.20 & 39.00 & \textbf{40.40} & \textbf{40.20} & 39.40 & 39.60 & \textbf{40.20} & 38.60  \\
BoolQ       & \textbf{67.22} & 66.48 & 67.22 & 64.43 & \textbf{68.20} & 67.22 & \textbf{67.55} & 66.85 & \textbf{67.22} & 65.44 \\
PIQA        & \textbf{76.82} & \textbf{76.82} & \textbf{76.82} & 76.44 & 76.39 & \textbf{76.82} & 76.44 & 76.55 & \textbf{76.82} & 76.39 \\
HellaSwag   & \textbf{67.97} & 67.88 & 67.97 & 67.80 & \textbf{68.01} & \textbf{67.97} & \textbf{67.97} & 67.93 & \textbf{67.97} & 66.93 \\
\bottomrule
\end{tabular}}
\label{tab:Ablation}
\end{table*}

\noindent{\textbf{LLMs and Benchmarks.}}
To demonstrate how QPruner performes on different model, we test it on three open source large language models: LLaMA-7B \cite{touvron2023llama}, LLaMA-13B \cite{touvron2023llama} and Vicuna-7B \cite{zheng2024judging}, and specific version is stated in the Appendix \ref{LLM}. We conduct these LLMs on zero-shot classification tests for commonsense reasoning datasets, including BoolQ \cite{clark2019boolq}, PIQA \cite{bisk2020piqa}, HellaSwag \cite{zellers2019hellaswag}, WinoGrande \cite{sakaguchi2021winogrande}, ARC-easy \cite{clark2018think}, ARC-challenge \cite{clark2018think}, and OpenbookQA \cite{mihaylov2018can}.

\noindent{\textbf{Software and hardware configuration.}}
We utilize the following configurations: \textit{PyTorch} version 2.1.2, \textit{BitsandBytes} library version 0.43.1, \textit{Transformers} library version 4.41.0, \textit{PEFT (Parameter-Efficient Fine-Tuning)} library version 0.11.1, \textit{Optuna} library version 3.6.1, \textit{CUDA} version 12.4, \textit{GPU:} NVIDIA L20 GPU with 48GB of memory.

\noindent \textbf{Implementation Details.}
The pruning method follows LLM-Pruner \citep{ma2023llm}, and the dataset uses 50k publicly available samples from the Alpaca \cite{alpaca}. All experiments were conducted with a LoRA matrix rank of 8, and LoftQ initialization with one iteration. We utilized BitsandBytes for quantization configuration, for memory considerations, we keep the number of 8-bit layers below 25\%. For 4-bit quantization, we employed NF4 \cite{dettmers2024qlora}, and since 2-bit quantization does not reduce memory usage, each layer's quantization configuration only considered 4-bit and 8-bit options. More detailed hyperparameter settings can be found in Appendix \ref{appendix:hyperparams}.

\subsection{Main Results}
\label{MainResults}
In this section, we present experimental results to demonstrate the capability of our proposed QPruner framework in balancing performance while reducing memory usage through integrating quantization and structured pruning. Through further iterative optimization, it can even achieve better performance than high-precision models. Although pruning methods are very important, the pruning method itself is not our focus; therefore, we adopt the popular LLM-Pruner~\cite{ma2023llm} as our baseline, which is a widely used structured pruning method that directly removes weights.

We evaluate the model performance and peak memory usage of LLM-Pruner and QPruner under different pruning rates. Due to the lack of specific test prompts in the LLaMA paper, we utilize open-source prompts provided by \citet{eval-harness} for benchmarking. Results for the LLaMA-7B and Vicuna-7B models are shown in Table~\ref{table:result-comp}, and results for LLaMA-13B are provided in Appendix~\ref{res-llama13b}. Although our method is expected to have greater advantages on larger models (e.g., 70B parameters or more), due to hardware limitations, we focus only on models within 13B parameters.

In our experiments, \textbf{QPruner\textsuperscript{1}} denotes the use of uniform quantization across all layers, \textbf{QPruner\textsuperscript{2}} represents the mixed-precision configuration based on mutual information, and \textbf{QPruner\textsuperscript{3}} refers to the mixed-precision quantization after further optimization using Bayesian methods based on QPruner\textsuperscript{2}. Theoretically, full-parameter fine-tuning would perform better than PEFT methods; however, it performs poorly on the Alpaca dataset commonly used in model compression. If we perform individual training according to each benchmark, only the pruned models after quantization can be fully fine-tuned, which is an advantage of our framework, but this would lead to unfair comparisons. Therefore, for unquantized models, we use LoRA~\cite{hu2021lora} fine-tuning, and for quantized models, we use LoftQ~\cite{li2023loftq} fine-tuning.

From Table~\ref{table:result-comp}, we observe that our method demonstrates more significant advantages at higher pruning rates. For instance, at a pruning rate of 50\% on the LLaMA-7B model, \textbf{QPruner\textsuperscript{3}} outperforms LLM-Pruner by achieving a higher accuracy on the BoolQ dataset (48.37\% vs.\ 43.76\%) while reducing memory usage from 23.89\,GB to 16.65\,GB—a reduction of approximately 30\%. This highlights the effectiveness of our framework in maintaining or even improving performance under aggressive compression.

These results demonstrate that our QPruner framework effectively balances memory efficiency and model accuracy by integrating quantization with structured pruning. By employing finer-grained quantization strategies and a combined performance recovery phase, we mitigate the detrimental effects that pruning and quantization individually impose on LLMs. This integration not only reduces memory consumption but can also enhance model performance, especially at higher pruning rates.

\subsection{Ablation Study}

We conducted ablation experiments using LLaMA-7B with a 20\% pruning rate, based on results obtained by \textbf{QPruner\textsuperscript{3}}. All results are presented in Table~\ref{tab:Ablation}. We tested different quantization data types (\texttt{NF4}, \texttt{FP4}), LoRA matrix initialization methods (Gaussian, PiSSA~\cite{meng2024pissa}, LoftQ), varying iteration counts in LoftQ (more iterations represent better error fitting), and different importance estimation methods.

Our experiments show that the choice of quantization data type slightly affects performance, but our method is effective across different types. Similarly, different LoRA initialization methods yield comparable results, indicating robustness to initialization strategies. Interestingly, increasing the number of iterations in LoftQ does not necessarily improve performance, suggesting that fitting residuals with low-rank matrices may not always be beneficial. Finally, using first-order Taylor approximations for importance estimation outperforms second-order ones, highlighting the complexity of LLMs and the limitations of higher-order approximations.

Additional experiments on different Bayesian optimization iteration counts and resource consumption are provided in Appendix~\ref{op}. The Pareto frontier demonstrates that more iterations can lead to better configurations, albeit at increased computational cost.
\section{Related Work}
\subsection{Efficient Compression of LLMs}
LLM-Pruner \cite{ma2023llm} uses structured pruning to eliminate non-essential interconnected structures by leveraging gradient information. This technique enables compressed models to maintain good performance across multiple tasks with basic fine-tuning. \citet{santacroce2023matters} proposes Globally Unique Movement (GUM), a novel pruning technique focusing on the sensitivity and uniqueness of LLMs' network components. GUM prunes neurons that uniquely contribute to the model output and are sensitive to loss changes, thus preserving high accuracy. This method optimizes the trade-off between information retention and computational efficiency. Quantization-Aware Training (QAT) combines quantization with full model fine-tuning to adapt models for downstream tasks \citep{peri2020deploying, liu2023llm}. Although QAT is effective, it requires substantial computational resources, such as gradient calculations and optimization states, and it complicates the gradient computation for quantized weights. However, by leveraging LoRA, these challenges can be bypassed during task adaptation. Post-Training Quantization (PTQ) frameworks, such as GPTQ and SmoothQuant \citep{frantar2022gptq, xiao2023smoothquant}, use a small subset of training data to calibrate high-precision models, enabling the generation of task-specific quantized models without the need for gradient backpropagation. This makes PTQ more cost-efficient than QAT, although it generally results in lower accuracy. \citet{xiao2023smoothquant} proposed SmoothQuant, a post-training quantization framework that employs a mixed-precision strategy to calibrate large language models, enabling accurate and efficient deployment without the need for retraining.

\subsection{Parameter Efficient Fine-Tuning}
LLM-Adapters \cite{hu2023llm} integrate small adapters with few extra parameters into LLMs for efficient fine-tuning, allowing smaller models to perform as well as larger ones on specific tasks. Unlike the serial approach of adapters, low-rank adaptation (LoRA) \cite{hu2021lora} uses a parallel method to insert trainable rank decomposition matrices into each layer of the model's architecture. LoRA adds trainable matrices to each layer while keeping the pre-trained weights unchanged, reducing the number of trainable parameters and making model adaptation faster and less resource-intensive. QLoRA \cite{dettmers2024qlora} combines low-rank adapters and quantized 4-bit weights for efficient LLM fine-tuning, significantly reducing GPU memory requirements while achieving performance comparable to full 16-bit fine-tuning. LoftQ \cite{li2023loftq} applies quantization and low-rank approximation alternately to achieve a good initialization for LoRA fine-tuning, mitigating the discrepancy between quantized and pre-trained weights, and enabling efficient fine-tuning of quantized models, particularly in challenging low-bit regimes.

\section{Conclusion}
We propose QPruner, an innovative framework that combines structured pruning and quantization for efficient model compression. Given that structured pruning and quantization typically require performance recovery steps, integrating them provides a more holistic approach to mitigating the errors introduced by both techniques while further compressing the model. To address the uneven importance distribution across layers and precision loss caused by pruning and quantization, we adopt a fine-grained method to preserve the capacity of critical layers, enhancing their performance further during the fine-tuning process. After pruning, we first allocate mixed-precision quantization based on task relevance, followed by Bayesian optimization to iteratively refine decisions and probabilistically select the optimal quantization configuration. Experimental results demonstrate that QPruner significantly outperforms baseline models in terms of memory efficiency while achieving superior accuracy across multiple NLP benchmarks. By striking a balance between efficiency and performance, shows that QPruner is a powerful solution for deploying LLM in resource-limited environments.

\section*{Limitation}
One of the current limitations of QPruner is the significant precision loss caused by structured pruning, which still impacts the overall model performance. In future work, we aim to further optimize the pruning process to minimize this precision degradation. Additionally, the use of Bayesian optimization requires real data to guide the process, which can be time-consuming. While this method improves quantization configurations, the iterative nature of Bayesian optimization introduces additional computational overhead that may not be ideal for all deployment scenarios.
\bibliography{custom}

\clearpage
\appendix
\section{Version of LLMs}\label{LLM}
We provide the Hugging Face link of LLMs used in the experiment:
LLaMA-7B: \url{https://huggingface.co/baffo32/decapoda-research-llama-7B-hf}; Vicuna-7B: \url{https://huggingface.co/lmsys/vicuna-7b-v1.5}; LLaMA-13B: \url{https://huggingface.co/yahma/llama-13b-hf}

\section{Hyperparameters}
\label{appendix:hyperparams}
In the optimization of the pruned LLaMA-7B model, a comprehensive hyperparameter configuration was employed to ensure an optimal balance between model performance and computational efficiency. The model was fine-tuned with a learning rate of $3 \times 10^{-4}$, utilizing a batch size of 8, further divided into micro batches of 4 to manage memory constraints effectively. Sequences were standardized to a maximum length of 256 tokens, and a dropout of 0.05 was applied specifically to the LoRA layers targeting projections such as query, key, value, and output, alongside gate, down, and up projections. Quantization was dynamically applied at 4-bit and 8-bit levels according to layer requirements to optimize memory use without compromising computational accuracy. The training employed the paged AdamW optimizer with 32-bit precision, enhancing stability and efficiency. These settings were methodically tested and optimized through the Optuna framework to ensure robust model performance and resource utilization.

\section{Results of Optimization Workflow}\label{op}
\label{appendix:pareto}
\begin{figure*}[h]
    \centering
    \begin{subfigure}{0.45\linewidth}
        \centering
        \includegraphics[width=\linewidth]{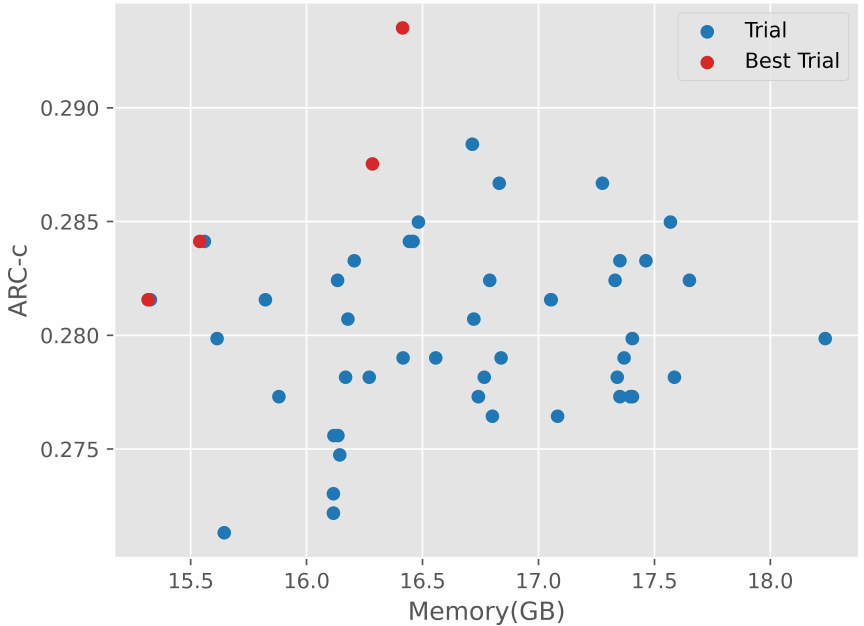}
        \caption{ARC-c}
        \label{fig:sub1}
    \end{subfigure}
    \begin{subfigure}{0.45\linewidth}
        \centering
        \includegraphics[width=\linewidth]{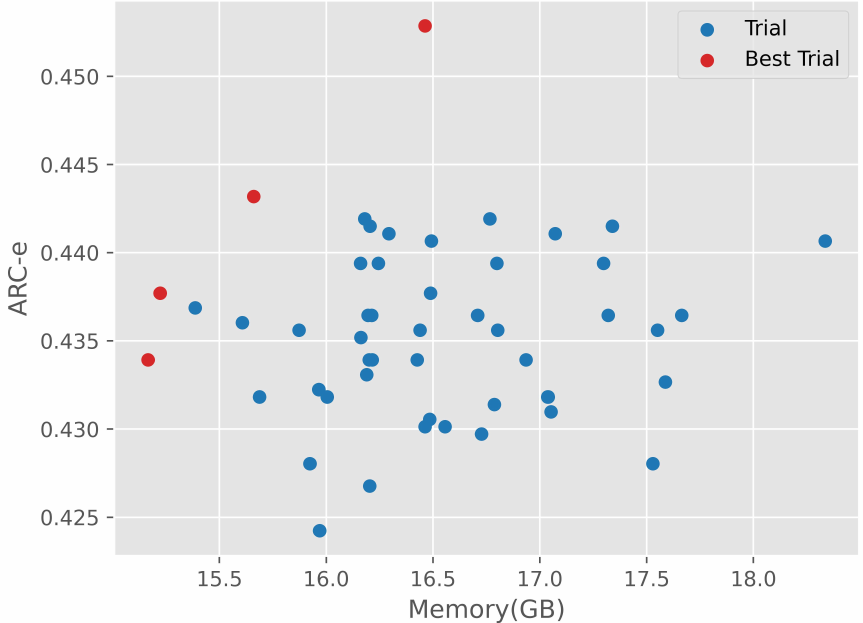}
        \caption{ARC-e}
        \label{fig:sub2}
    \end{subfigure}
    \begin{subfigure}[b]{0.45\linewidth}
        \centering
        \includegraphics[width=\linewidth]{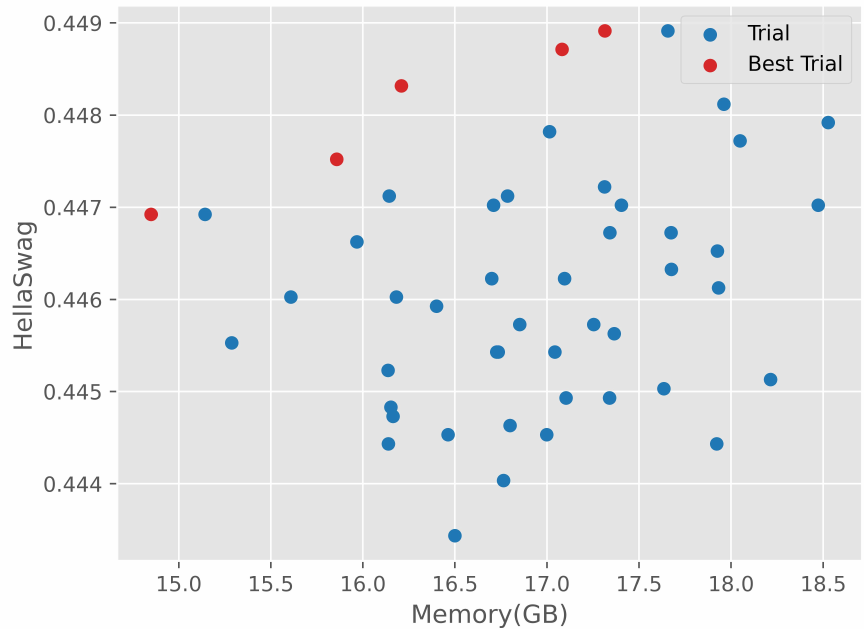}
        \caption{HellaSwag}
        \label{fig:sub3}
    \end{subfigure}
    \begin{subfigure}[b]{0.45\textwidth}
        \centering
        \includegraphics[width=\linewidth]{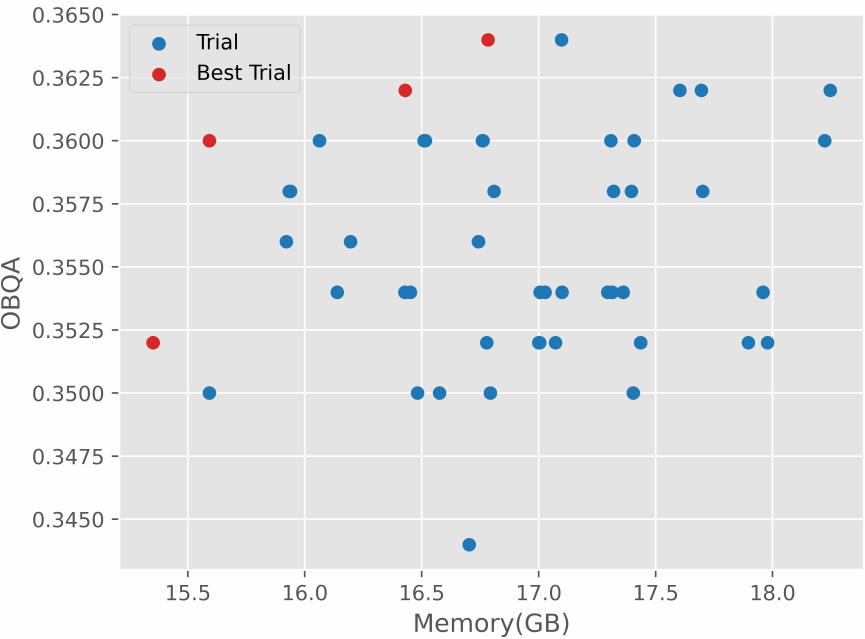}
        \caption{OBQA}
        \label{fig:sub4}
    \end{subfigure}
    \begin{subfigure}[b]{0.45\textwidth}
        \centering
        \includegraphics[width=\linewidth]{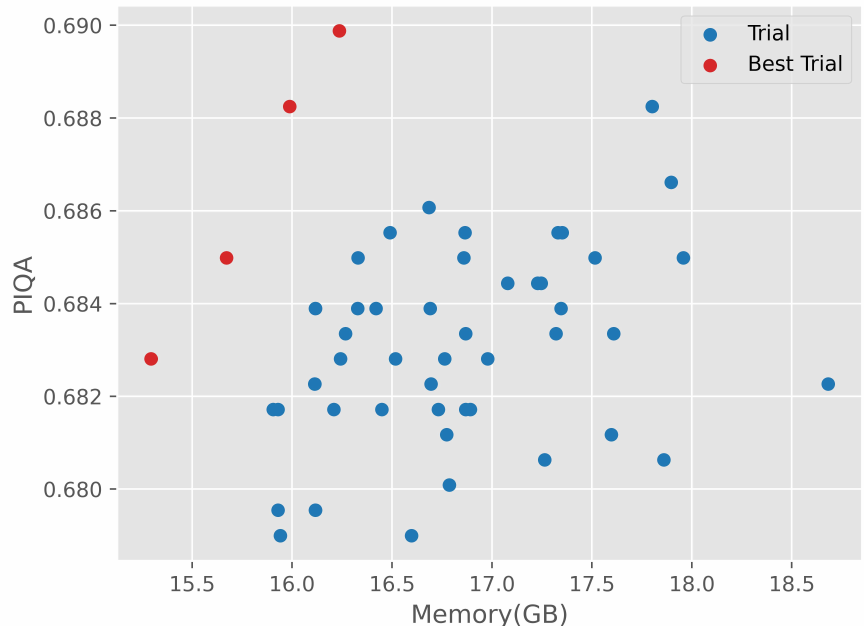}
        \caption{PIQA}
        \label{fig:sub5}
    \end{subfigure}
    
    \caption{Pareto-front scatter plots for different Downstream Tasks}
    \label{fig:combined_pareto_fronts}
\end{figure*}

In this section, we will use the LLaMA-7B model with 50\% pruning as our example to illustrate the Pareto optimization workflow, as shown in Figure \ref{fig:combined_pareto_fronts}

\section{\textbf{Details of The Optimization Workflow.}}
\label{pareto_train}

\begin{figure}[h]
    \centering
    \begin{subfigure}[b]{0.45\textwidth}
        \centering
        \includegraphics[width=\textwidth]{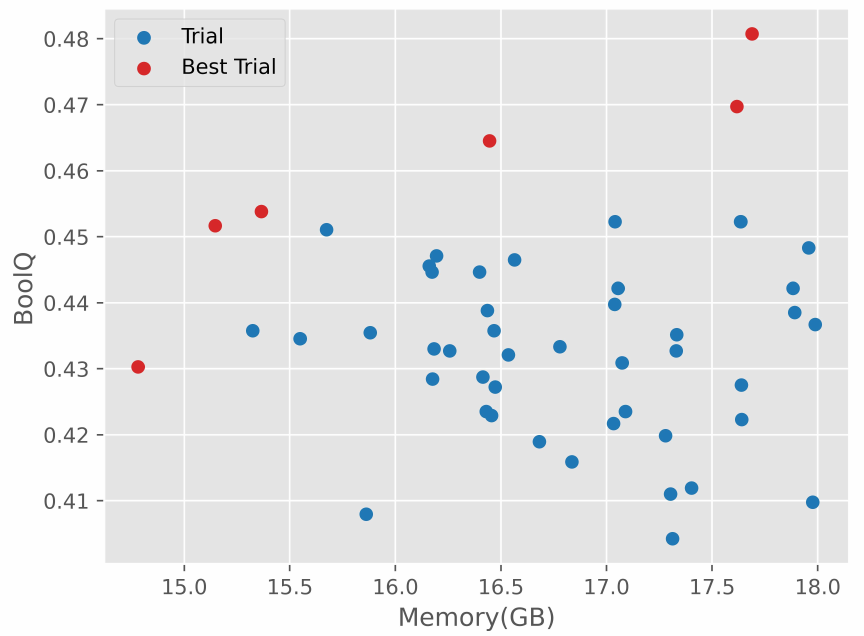}
        \caption{Pareto-front scatter plot for BoolQ}
        \label{fig:sub1}
    \end{subfigure}
    \hfill 
    \begin{subfigure}[b]{0.45\textwidth}
        \centering
        \includegraphics[width=\textwidth]{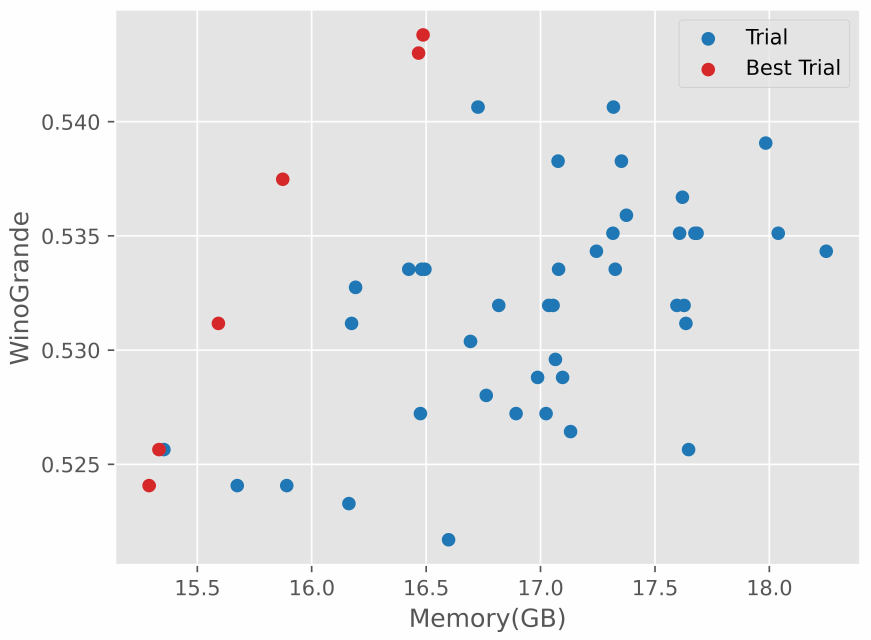}
        \caption{Pareto-front scatter plot for WinoGrande}
        \label{fig:sub2}
    \end{subfigure}
    \caption{Pareto-front scatter plots for BoolQ and WinoGrande with 50 data points. The red points indicate the non-dominated configurations within the Pareto frontier.}
    \label{fig:test}
\end{figure}

We illustrated the optimization process, memory, and time footprint of QPruner using a 50\% parameter pruning rate on llama-7b as an example. We fine-tuned 10 sets of configurations as the initialization for the Gaussian Process (GP) (this is not mandatory; in other experiments, we found that starting from scratch, a good configuration could be found in about 10 iterations). In each configuration, the quantization precision for all model layers was randomly selected between 4-bit and 8-bit. On average, obtaining data for each initialization took approximately 25 minutes. We set the total number of iterations for QPruner to 40 (resulting in 50 data points for constructing the Pareto front) to ensure the best configuration was found. The entire process took approximately 16.5 hours. During QPruner iterations, GP required around 7s to suggest the next configuration, while the prediction process and Pareto frontier construction consumed approximately 187MB memory. In Figure \ref{fig:test}, we present the optimization results for BoolQ and Winograd. More detailed processes and results for other benchmarks are provided in Appendix \ref{appendix:pareto}.

\section{Performance in LLaMA-13B}\label{res-llama13b}
We list the performance of the configuration described in Section \ref{4.1} for LLaMA-13B in Table \ref{llama13bperform}.

\section{Code Usage Instructions}
In our workflow, we begin by defining a custom pruning rate to prune the original LLM model. This step generates a new, streamlined model version that we save under the “tuning” directory. Subsequently, we apply either random or specified precision quantization adjustments to this pruned model. These quantization strategies are carefully chosen to further reduce the model’s memory footprint while striving to maintain its operational effectiveness. Once quantized, the model undergoes a thorough evaluation using specific tools that assess its performance on downstream tasks to ensure it retains accuracy and effectiveness after modifications.

To integrate Optuna into our optimization process, we begin by recording the pruning and quantization parameters in a JSON format. These parameters serve as historical data inputs for Optuna, facilitating a structured approach to hyperparameter tuning. We employ Optuna to conduct multi-objective optimization, setting multiple goals such as balancing model performance with memory efficiency. Optuna's iterative search process explores various parameter combinations across numerous iterations, each informed by the outcomes of previous evaluations. The results from these searches are compiled into a dataframe, which not only aids in subsequent analysis but is also crucial for identifying the Pareto frontier, optimizing the trade-offs between performance and resource usage.

\section{Limitations}
This study focuses solely on the fine-tuning of pruned models, leveraging the self-adjusting nature of QPruner. However, we believe that this adaptive approach could be broadly applicable to a wide range of models, not limited to pruned ones. Future research will explore the optimization of fine-tuning across different model architectures.

Additionally, we did not conduct experiments on larger models, such as those with 70B parameters or more. The scalability and effectiveness of QPruner on such massive models remain to be investigated, and this will be a key focus of our future work.
\begin{table*}[h]
    \centering
    \resizebox{\textwidth}{!}{
        \begin{tabular}{llccccccc}
            \toprule
            Pruning Rate & MethodQPruner\textsuperscript{1} & BoolQ & PIQA & HellaSwag & WinoGrande & ARC-e & ARC-c & OBQA \\
            \midrule
            {\parbox{1.8cm}{Rate = 0\%}} & w/o tuning & 68.50 & 79.11 & 76.21 & 70.09 & 74.58 & 44.54 & 42.20  \\
            \cmidrule{1-9}
            \multirow{3}{*}{\parbox{1.8cm}{Rate = 50\%}} 
            & LLM-Pruner     & 61.93(41.32) & 71.38(41.32) & 53.36(41.32) & 53.59(41.32) & 29.95(41.32) & 53.11(41.32) & 38.00(41.32)  \\
            \cmidrule{2-9}
            & QPruner\textsuperscript{1}    & 61.71(36.68) & 72.63(36.68) & 56.10(36.68) & 55.17(36.68) & 31.57(36.68) & 55.47(36.68) & 38.60(36.68)  \\
            \cmidrule{2-9}
            & QPruner\textsuperscript{3}     & 61.80(30.53) & \textbf{73.23}(30.53) & \textbf{56.37}(30.53) & 55.09(31.45) & 31.48(30.53) & \textbf{55.80}(31.45) & \textbf{39.00}(30.58)  \\
            \bottomrule
        \end{tabular}}  
    \caption{Zero-shot performance and memory. ‘Bold’ indicates the best performance at each pruning rate.  Reported in percentage (\%).}
    \label{llama13bperform}
\end{table*}

\end{document}